\documentclass[10pt,twocolumn,letterpaper]{article}

\usepackage{cvpr}      
\usepackage{algorithm}
\usepackage{algpseudocode}
\usepackage{tabularx}   

\usepackage{capt-of} 

\usepackage{stfloats} 

\definecolor{cvprblue}{rgb}{0.21,0.49,0.74}
\usepackage[pagebackref,breaklinks,colorlinks,allcolors=cvprblue]{hyperref}

\def\paperID{*****} 
\def\confName{CVPR}
\def\confYear{2025}

\title{PLLM: Pseudo-Labeling Large Language Models for CAD Program Synthesis}

\author{
Yuanbo Li\\
Brown University\\
Toyota Research Institute
\and
Dule Shu\\
Toyota Research Institute
\and
Yanying Chen\\
Toyota Research Institute
\and
Matt Klenk\\
Toyota Research Institute
\and
Daniel Ritchie\\
Brown University
}

\begin{document}
    \maketitle


\begin{abstract}
Recovering Computer-Aided Design (CAD) programs from 3D geometries is a widely studied problem. Recent advances in large language models (LLMs) have enabled progress in CAD program synthesis, but existing methods rely on supervised training with paired shape–program data, which is often unavailable.

We introduce PLLM, a self-training framework for CAD program synthesis from unlabeled 3D shapes. Given a pre-trained CAD-capable LLM and a shape dataset, PLLM iteratively samples candidate programs, selects high-fidelity executions, and augments programs to construct synthetic program–shape pairs for fine-tuning. We experiment on adapting CAD-Recode from DeepCAD to the unlabeled ABC dataset show consistent improvements in geometric fidelity and program diversity.\end{abstract}

\section{Introduction}
\label{sec:intro}

Computer-Aided Design (CAD) is the industry standard for 3D modeling in engineering and manufacturing. Designers typically construct models through a sequence of parametric operations, which, when executed, produce boundary representations (B-reps) of 3D geometry. The inverse problem of recovering a CAD program from a given shape is also extensively studied. Recovering the program enables semantic editing, programmatic modification, and compact representation of 3D models.

Previous approaches address this inverse problem by training lightweight neural networks to predict CAD operations and their corresponding parameters ~\cite{Benko2001ReverseBRep, Dupont2022CADOpsNet, Zhou2023CADParser, ZoneGraphs, Ali2024BRepDetNet}. More recently, large language models (LLMs) have been explored for this reverse-engineering task due to their strong symbolic reasoning abilities and rapid progress in program synthesis ~\cite{Rukhovich2025CADRecode, Wang2025CADFusion, wang2024cadgpt, Mallis2024CADAssistant, Li2025CADLlama}. However, existing methods depend on supervised training with ground-truth CAD programs. Such annotations are expensive, dataset-specific, and often unavailable, creating a major bottleneck for scaling CAD program synthesis to new domains and new CAD languages.

In this work, we introduce a new framework to address this the lack of data. Rather than relying on labeled programs, we derive supervision from the data generation process itself. Formally, we assume a pre-trained LLM $p(z \,|\, x, \mathcal{L})$ that generates a CAD program $z$ in language $\mathcal{L}$ from a shape $x \sim \mathcal{S}$. Given a new shape distribution $\mathcal{S}^*$ without program annotations, our goal is to adapt the model by automatically generating informative training data from $\mathcal{S}^*$ itself.

Our observation is that the pre-trained LLM already encodes substantial knowledge about CAD structure and can produce plausible programs even for unseen domains, though they might be suboptimal. We therefore treat model outputs not as final predictions but as starting point for data synthesis. Our framework samples programs, executes them, compares the resulting geometry to the input shape, and selectively retains high-fidelity results. We further expand this data by programmatic edits and variations, producing enriched program–shape pairs that provide stronger supervision than the original predictions alone. Through iterative self-training, the model gradually improves using this growing synthetic dataset.

Specifically, we instantiate this framework using CAD-Recode ~\cite{Rukhovich2025CADRecode}, trained on DeepCAD~\cite{DeepCAD} to generate programs in the CadQuery language. We then adapt it to the ABC dataset~\cite{ABCDataset}, a widely used shape collection that lacks ground-truth CAD programs. This setting highlights the ability of our approach to synthesize useful supervision where none exists.

In summary, we propose a novel method to fine-tune existing LLMs for improved CAD program synthesis for new domain in the absence of ground-truth supervision. Our contributions are as follows:

\begin{itemize}
    \item We introduce PLLM, a self-training framework that synthesizes and enriches CAD programs to construct supervision for unlabeled 3D datasets.
   
    \item We propose a data synthesis strategy that expands model outputs through sampling and programmatic edits, generating diverse and informative program–shape pairs.

    \item We demonstrate that synthetic self-training improves geometric fidelity when adapting CAD-Recode from DeepCAD to the unlabeled ABC dataset.
\end{itemize}

\section{Related Works}
\label{sec:related_works}
\subsection{Self Supervised Training}
Our work lies in the broader area of unsupervised and weakly-supervised learning~\cite{williams1992simple,bunel2018leveraging}, where the central challenge is how to obtain useful supervision when labeled data is scarce or unavailable. A common solution in such settings is reinforcement learning (RL)~\cite{sutton2000policy,williams1992simple,ziebart2008maximum,silver2014deterministic,mnih2016asynchronous}, which treats supervision as a reward signal. However, CAD programs involve discrete, non-differentiable operations and expensive execution, making RL-based optimization unstable and inefficient for our setting. Instead, we adopt a self-training paradigm, where the model generates its own supervision. Self-training has a long history in weakly-supervised learning~\cite{mcclosky2006effective,scudder1965probability,yarowsky1995unsupervised}, and recent work shows that combining self-training with data augmentation and synthetic data generation can substantially improve neural models across domains~\cite{he2020revisiting,kahn2020self,zoph2020rethinking}. Our approach builds on this perspective by treating model outputs as a source of synthetic data that can be filtered, expanded, and reused for training.

In visual program synthesis, self-training has recently emerged as an effective strategy for learning without ground-truth programs~\cite{jones2024editing, ganeshan2023coref, jones2023shapecoder}. Our method is closely related to execution-guided learning~\cite{chen2019execution,ellis2019write,ellis2018learning,ellis2020dreamcoder}, where program execution provides feedback on correctness. In our case, execution serves as a mechanism to evaluate and curate synthetic programs, allowing us to retain high-quality program–shape pairs as supervision.

PLAD~\cite{Plad} proposes a bootstrapped learning framework in which a pre-trained generator produces candidate programs that are then used to fine-tune the model. Our approach follows a similar high-level philosophy but extends it toward richer data synthesis: rather than only selecting successful generations, we further expand and refine programs through programmatic edits and variations. This transforms initial predictions into a growing and increasingly informative synthetic dataset. We instantiate this idea in the context of CAD program synthesis using a pre-trained LLM as the generator.

\subsection{Learning to Recover CAD Programs}
Our work also relates to the broader goal of reverse CAD engineering from diverse input modalities, including voxel grids~\cite{Sharma_2018_CVPR,ShapePrograms,LambournePrismatic}, point clouds~\cite{Wu2018, TaoInverseCSG,DeepCAD, liu2024point2cad, GuoComplexGen2022, sharma2020parsenet, liu2024point2cad, uy-point2cyl-cvpr22}, and boundary representations~\cite{ZoneGraphs}. Early approaches relied on heuristics or lightweight neural networks, while more recent works explore large language models due to their symbolic reasoning and program synthesis capabilities~\cite{Badagabettu2024Query2CAD, Wu2023CADLLM,Zhu2024Text2CAD,Alrashedy2025CADCodeVerify,Ocker2025FromIdea2CAD,Wang2025CADGPT, Guan2025CADCoder, Zhang2025LLMsCAD, Alrashedy2024CADCode, Makatura2023LLMsDesignManufacturing}. Our method belongs to this family of reverse CAD approaches.

However,a common limitation for existing methods~\cite{Xu2025CADMLLM, Rukhovich2025CADRecode,Khan2024VLMforCADFeatures, Li2025CADLlama, Khan2024CADSIGNet, Wang2025CADFusion} is that they rely on datasets containing paired ground-truth CAD programs and shapes. Such paired data is expensive to obtain and scarce in practice. Many large-scale CAD repositories~\cite{ABCDataset,mcm2023,simjeb2023,willis2021fusion} provide high-quality geometry but lack program annotations. This data imbalance limits the scalability of supervised reverse CAD systems.

Our approach addresses this challenge from a data synthesis perspective. Rather than requiring paired supervision, we treat unlabeled CAD collections as raw material for generating synthetic program–shape pairs. By leveraging a pre-trained model (CAD-Recode~\cite{Rukhovich2025CADRecode}) to propose programs and then expanding, filtering, and refining these programs through execution feedback and programmatic edits, we construct a growing synthetic dataset that enables adaptation to new domains without manual annotations.

\section{Method}
\label{sec:method}

In this section, we formally describe the PLLM framework, which treats CAD program synthesis as a process of iterative data construction and model refinement. The framework takes the following components as input:

\begin{itemize}
    \item \textbf{(1) Pre-trained LLM:} A model $p(z \,|\, x, \mathcal{L})$ capable of generating CAD programs $z$ from input shapes $x$ using the language $\mathcal{L}$, where $x$ is drawn from a source distribution $\mathcal{S}$.
    
    \item \textbf{(2) Target shape dataset:} A dataset of shapes $\mathcal{S}^*$ from a target distribution that differs from $\mathcal{S}$ and lacks program annotations.
    
    \item \textbf{(3) Black-box executor:} An executor $\mathcal{E}$ that executes a program $z$ to produce its corresponding 3D geometry.
\end{itemize}

Rather than assuming access to ground-truth programs, PLLM constructs supervision automatically from $\mathcal{S}^*$. The pre-trained model generates candidate programs for shapes in $\mathcal{S}^*$, which are executed and evaluated against the input shapes. High-quality executions are treated as synthetic supervision and accumulated into a training set that supports iterative refinement of the model.

The objective of PLLM is to adapt the pre-trained model to the new distribution $\mathcal{S}^*$ by leveraging this synthetic supervision. Through iterative self-training, we obtain an updated model $p'$. For an input shape $x^* \in \mathcal{S}^*$, a sampled program $z^* \sim p'(z \,|\, x^*, \mathcal{L})$ should execute to a geometry $\mathcal{E}(z^*)$ that better matches the input shape. We quantify this improvement using a geometric reward metric (Chamfer Distance), where lower distance indicates higher fidelity.

\begin{figure*}[t!]
  \includegraphics[width=\textwidth,height=5.1cm]{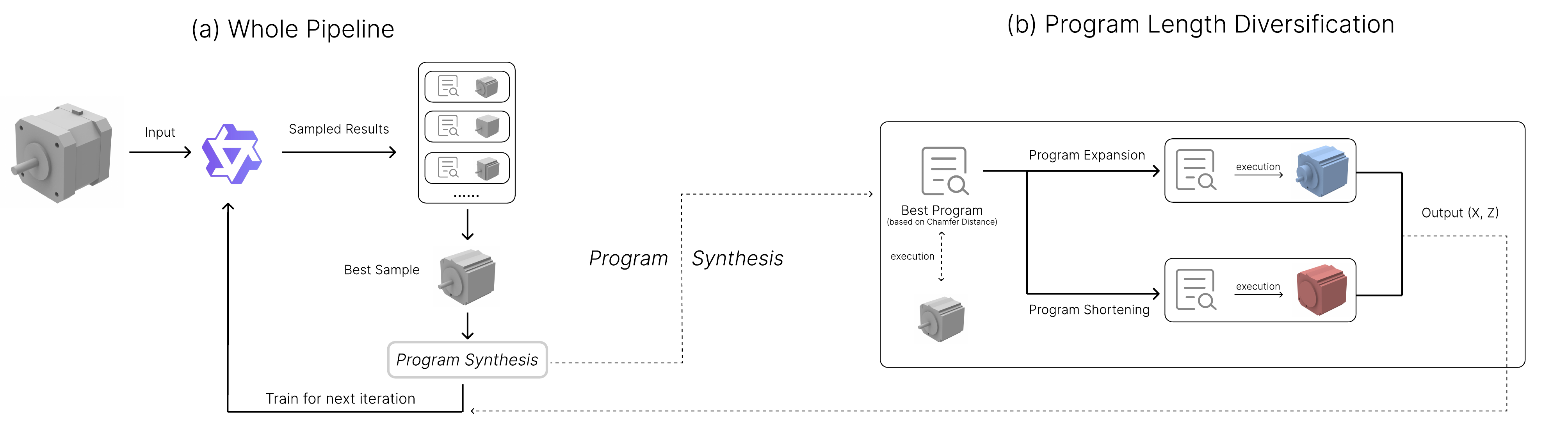}
  \caption{We show the overall pipeline in (a). At each iteration, the model first takes an input shape and samples multiple candidate programs. The selection algorithm then identifies the best program–shape pairs, which are used for training in the next iteration. (b) illustrates the details of the program length diversification process, where we perform both program expansion and shortening to create additional variants. The edited programs serve as labels $Z$, and their corresponding executions are treated as inputs $X$ to form the new training dataset.
}
  \label{approach_overview}
\end{figure*}

We illustrate the overall PLLM procedure in Figure~\ref{approach_overview}. PLLM adapts $p(z \,|\, x, \mathcal{L})$ to the target distribution $\mathcal{S}^*$ through an iterative data synthesis and self-training pipeline consisting of four key steps. First, the pre-trained model is used to sample multiple candidate programs for each input shape $x^* \in \mathcal{S}^*$ (Section~\ref{section_program_sampling}), producing diverse program hypotheses. Second, these candidates are executed and evaluated against the input shape, and the highest-fidelity programs are retained as initial synthetic supervision based on Chamfer Distance (Section~\ref{section_program_sampling}). Third, programmatic edits are applied to the selected programs to synthesize additional valid variants, enriching the program–shape pairs and exposing the model to a broader space of feasible solutions (Section~\ref{section_program_length_method}). Finally, the LLM is fine-tuned on this expanded synthetic dataset (Section~\ref{section_fine_tuning}).

Across iterations, these steps form a self-reinforcing data generation cycle: as the model improves, it produces higher-quality programs, which in turn yield more accurate and diverse synthetic training pairs. These improved pairs provide stronger supervision for subsequent updates, progressively aligning the model with the target distribution $\mathcal{S}^*$.

\subsection{Program Sampling}
\label{section_program_sampling}
Given an input shape $x^*$, the pre-trained LLM $p(z \,|\, x^*, \mathcal{L})$ generates $k=10$ candidate programs $\{z_i\}_{i=1}^k$ via stochastic decoding. We use nucleus sampling (top-$p=0.8$, top-$k=30$) with temperature 1.2 to encourage moderate diversity.

Each candidate is executed and compared to the input shape using Chamfer Distance. The program with the lowest distance is selected as the representative program $z^*$. If multiple candidates yield nearly identical reconstructions (difference $<10^{-4}$), we prefer shorter programs to promote concise representations.

\subsection{Program-Level Data Augmentation}
\label{section_program_length_method}
The pre-trained model may not fully capture the range of program structures required by the new distribution $\mathcal{S}^*$, particularly when shapes exhibit varying procedural complexity (see analysis in Section~\ref{section_eval_program_length}). As a result, directly using the model’s outputs can produce a narrow distribution of program lengths and structures.

To address this, we perform program-level data augmentation by synthetically expanding or shortening selected programs. 

\paragraph{Program Expansion}
We extend programs by appending additional operations to existing workspaces or by spawning new workspaces with procedurally generated sketch–feature sequences. We cap the total number of workspaces at $W_{\max}=5$ to maintain compact and executable programs while gradually increasing structural complexity.

\paragraph{Program Shortening}
We also generate shorter variants by removing top-level boolean operations (\texttt{union}, \texttt{cut}, \texttt{intersect}) while preserving syntactic validity. This produces more concise programs without drastically altering the resulting geometry.

These transformations create additional valid program–shape pairs that preserve geometric consistency while varying procedural structure. As a result, the synthetic dataset spans a broader range of program lengths and complexities, exposing the model to a wider spectrum of structural patterns and improving generalization to shapes with diverse complexity levels. This process is illustrated in Figure~\ref{approach_overview}(b).

\subsection{Training Data Pairs}
\label{section_fine_tuning}
We perform LoRA fine-tuning on the LLM using the synthetically constructed program–shape pairs, where both extended and shortened programs serve as $Z$ and their corresponding executions serve as $X$. A key advantage of this design is that each $(X, Z)$ pair is perfectly consistent: the shape $X$ is the direct execution result of program $Z$. This eliminates label noise and provides reliable supervision during fine-tuning.

By incorporating both extended and shortened programs, the synthetic dataset spans a broader range of program lengths and structural complexities. This enriched supervision improves the model’s ability to generalize across varying procedural patterns. In practice, this strategy maintains training stability while progressively expanding the model’s capacity to generate programs with diverse lengths and structures through iterative updates. We present additional experiments using alternative data-pair configurations in Section~\ref{section_plad_experiments}.

\section{Implementation}
\label{sec:implementation}

We use CAD-Recode~\cite{Rukhovich2025CADRecode}, pre-trained on DeepCAD~\cite{DeepCAD}, as our base model. The target domain $\mathcal{S}^*$ is the ABC dataset~\cite{ABCDataset}. Program execution is performed using CadQuery and its interpreter~\cite{cadquery}.

\subsection{CAD-Recode}

We adopt CAD-Recode~\cite{Rukhovich2025CADRecode} as our base model. CAD-Recode performs reverse CAD by mapping an input point cloud to executable CadQuery code. It consists of a point-cloud encoder that produces feature embeddings and a language-model decoder that generates CAD programs conditioned on these embeddings.

CAD-Recode is originally trained on the DeepCAD dataset using sketch–extrude programs. Since ABC shapes often require more diverse operations, we approximate them using sketch–extrude sequences rather than exact reconstruction. We also increase the maximum program length from 768 to 1200 tokens and apply our program diversification strategy to encourage longer and more detailed programs.

\subsection{LoRA Fine-Tuning}

We fine-tune CAD-Recode using Low-Rank Adaptation (LoRA) to support longer and more complex program generation. LoRA is applied to the middle transformer layers (layers 4–8) while keeping the remaining layers frozen. We use rank $r=8$, $\alpha=32$, and dropout $p=0.1$, applied to both self-attention and MLP projections.

\subsection{Computational Cost}

We use 75,000 shapes from the first 15 ABC batches (5,000 per batch). Of these, 71,784 yield executable programs and are used for experiments.

Training is performed on four NVIDIA L40S GPUs (48GB each) and an AMD EPYC 7R13 CPU. Six self-training iterations take ~150 hours total (~25 hours/iteration): ~12 hours for sampling, ~10 hours for execution and selection, and ~2 hours for four epochs of LoRA fine-tuning.

\section{Results and Evaluations}
\label{sec:experiments}
\begin{figure*}[t!]
  \includegraphics[width=\textwidth,height=3.75cm]{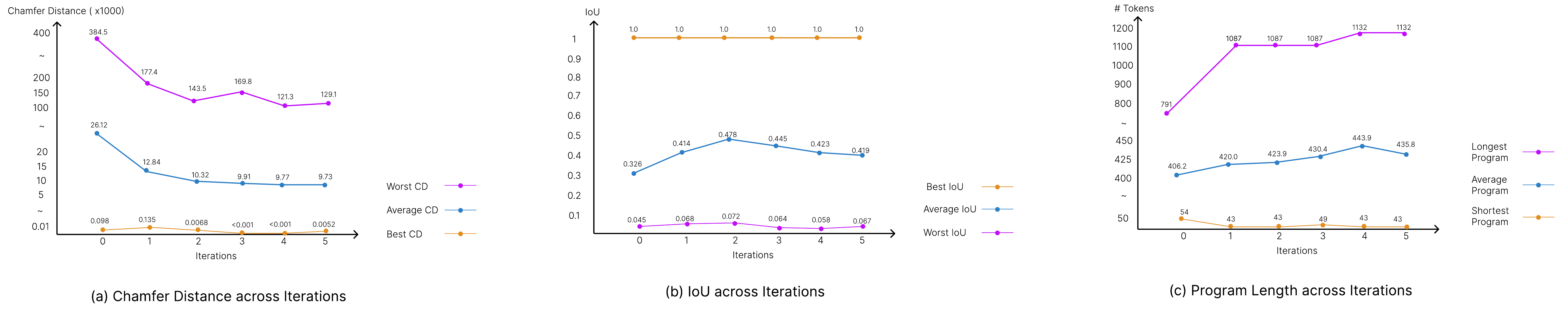}
\caption{We compare quantitative results across iterations: (a) Chamfer Distance, (b) IoU, and (c) Program Length.}

  \label{all_plot}
\end{figure*}

\begin{figure*}[t!]
  \includegraphics[width=\textwidth,height=5.9cm]{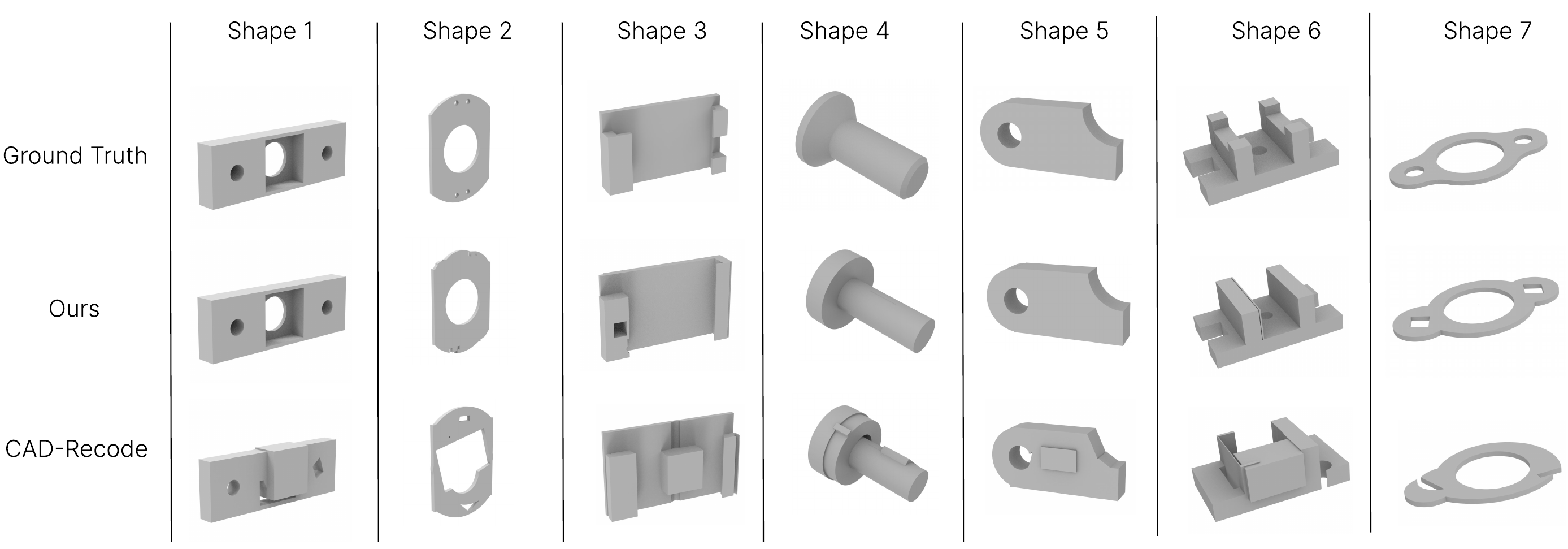}
  \caption{Comparison between our results and those produced by CAD-Recode, which correspond to the outputs from the first iteration of our framework}
  \label{fig_compare}
\end{figure*}

\begin{figure*}[t!]
  \includegraphics[width=\textwidth,height=13.2cm]{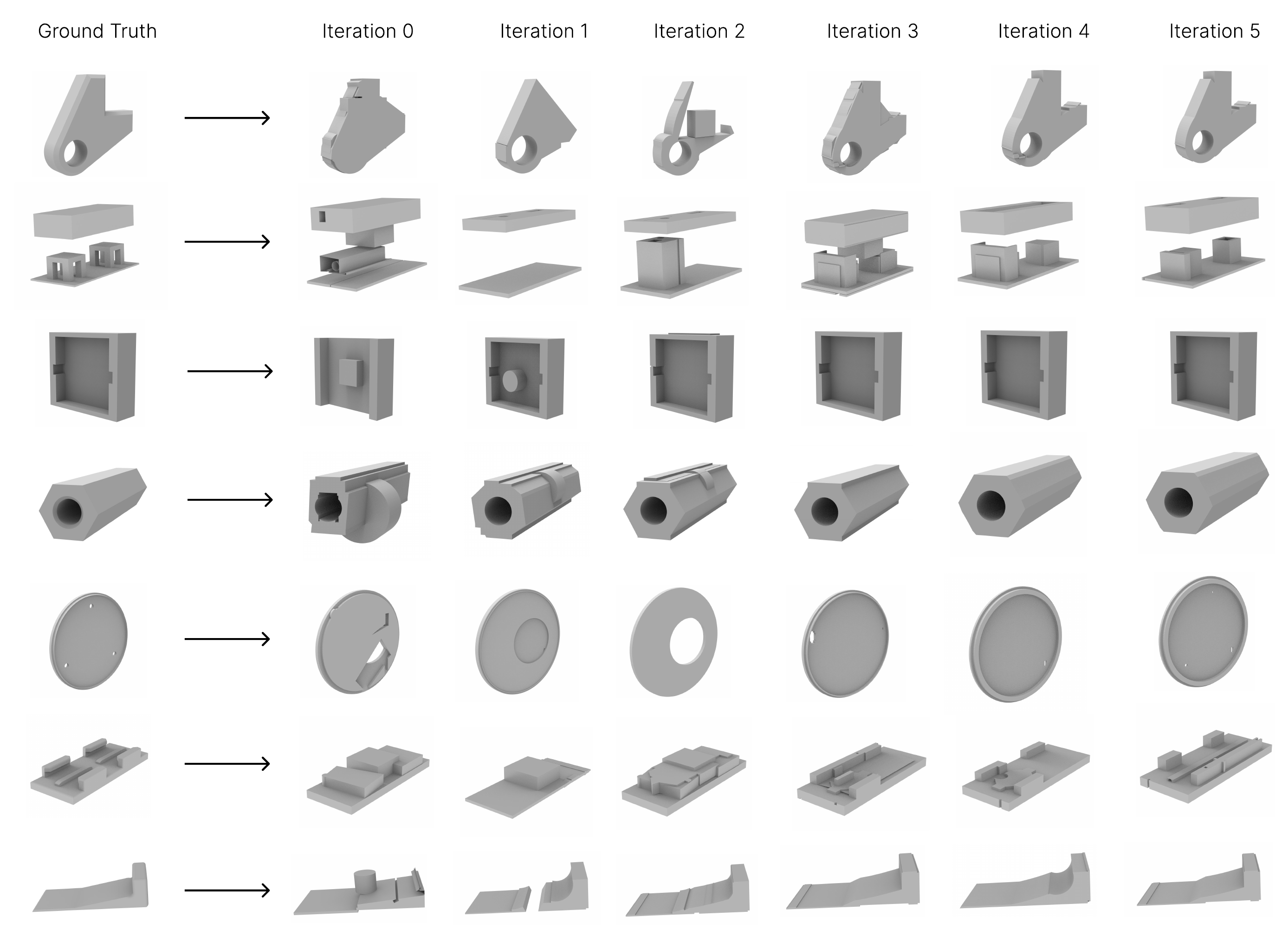}
  \caption{Results across different iterations, showing that the generated shapes gradually improve in quality as training progresses}
   \label{fig_across_it}
\end{figure*}

We sample point clouds from shapes in the ABC dataset and process them through the PLLM pipeline. We report qualitative comparisons against CAD-Recode (Figure~\ref{fig_compare}) and show results across training iterations (Figure~\ref{fig_across_it}). Quantitative evaluations include Chamfer Distance, Intersection over Union (IoU), and program length (Figure~\ref{all_plot}; Sections~\ref{section_CD}, \ref{section_IoU}, and \ref{section_eval_program_length}).

\subsection{Chamfer Distance Across Iterations}
\label{section_CD}
We report the best, average, and worst Chamfer Distances across iterations in Figure~\ref{all_plot}(a). Distances are computed after normalizing both predicted and input shapes to a unit bounding box ($1^3$) and scaling by $10^3$. The best and worst values are the mean Chamfer Distance of the top 10 and bottom 10 shapes per iteration, respectively, while the average is computed over all shapes. Chamfer Distance decreases consistently over the first four iterations, indicating improved geometric fidelity. Afterward, improvements plateau, which we attribute to the limited operation set of the base model CAD-Recode.

\subsection{IoU Across Iterations}
\label{section_IoU}
Another interesting metric to consider is the IoU across iterations (Figure~\ref{all_plot}(b)), which is not directly optimized in our framework. We do not intentionally select programs with high IoU, as our objective focuses on minimizing the Chamfer Distance (CD). While IoU measures volumetric overlap, CD evaluates surface alignment between the generated and target shapes. In our results, we observe that IoU increases during the first two iterations but decreases in later ones. This behavior arises because IoU is not explicitly used as a reward signal—thus, as the model focuses more on lowering CD, it may overfit surface alignment without necessarily improving volumetric consistency.

\subsection{Program Length Distance Across Iterations}
\label{section_eval_program_length}
We analyze how average, longest, and shortest program lengths evolve across iterations in Figure~\ref{all_plot}(c). Initially, average length increases, allowing finer shape generation. The baseline model, CAD-Recode, is limited to 768 tokens. When this cap is raised to 1200 tokens at iteration 0, program length grows slightly. From iteration 2 onward, as longer programs are added through expansion (see Section~\ref{section_program_length_method}), the maximum length rises markedly, improving the model’s capacity to represent detailed geometries.

\begin{figure}[h!]
\includegraphics[width=\linewidth]{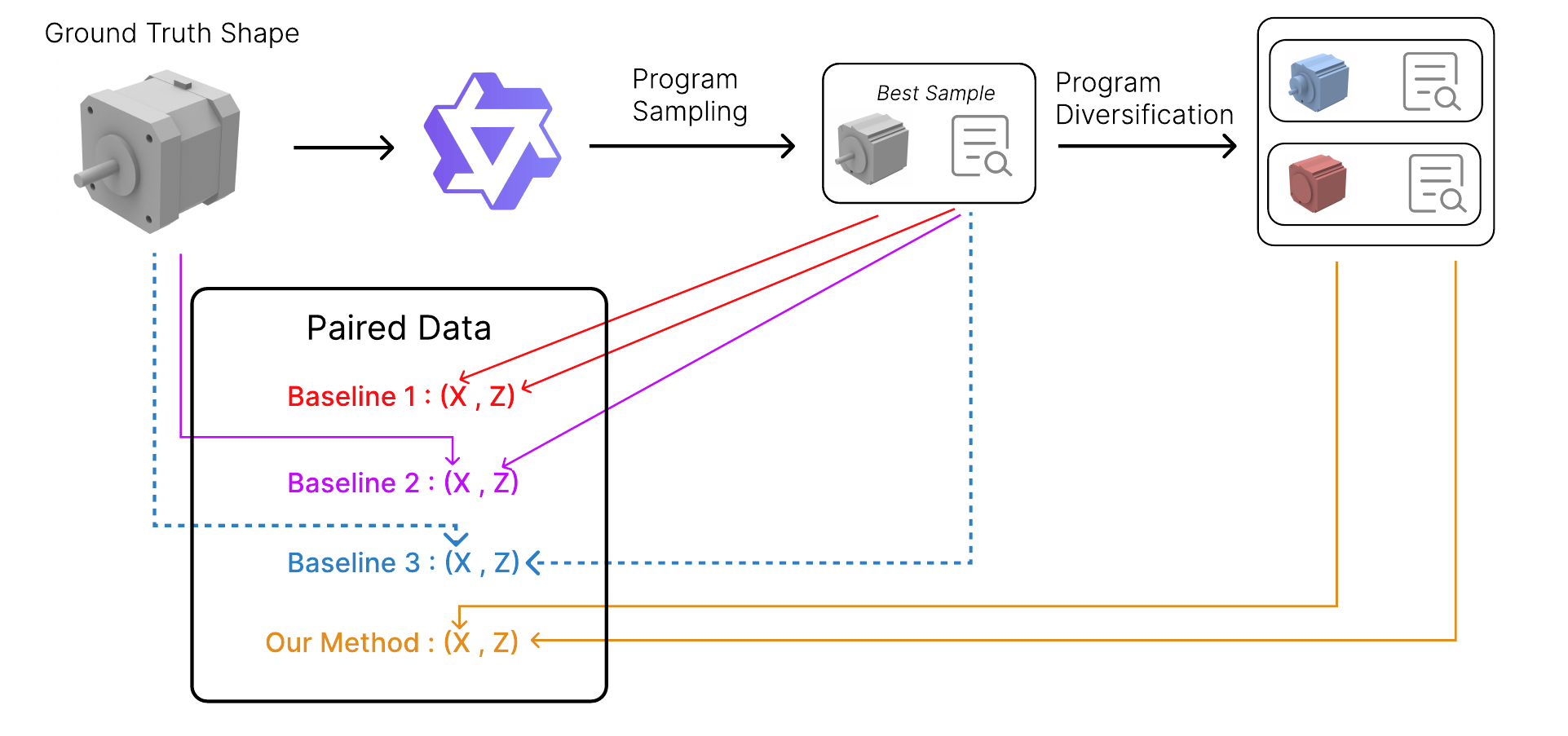}
\caption{Overview of different baseline strategies compared in our study. The figure illustrates how each baseline constructs its $(X, Z)$ training pairs. Baseline 1 uses the generated program and its execution; Baseline 2 uses the input shape and its best generated program; and Baseline 3 samples within each batch, selecting only the top 20\% of high-performing pairs. Our proposed method further introduces program expansion and shortening to generate paired data $(X, Z)$ that better align with the target distribution.}
   \label{fig_baseline_comparison}
\end{figure}

\subsection{Experiments with Different Pseudo Label Pairs}
\label{section_plad_experiments}

To fine-tune the model, pseudo program–shape pairs should ideally: (1) use high-quality programs, (2) provide executable program–shape consistency, (3) reflect the target shape distribution, and (4) introduce informative program variations. In practice, pseudo-labeling methods cannot satisfy all criteria simultaneously, leading to trade-offs.

Our approach selects top-performing programs (Criterion 1) and uses diversified programs with their executions for training, ensuring program–shape consistency and additional structural variation (Criteria 2 and 4). We partially satisfy Criterion 3 by anchoring supervision to executions derived from target-domain shapes.

We compare against alternative pairing strategies (Figure~\ref{fig_baseline_comparison}). Final-iteration results are summarized in Table~\ref{tab:different_pairs}, where our method achieves the best overall performance.

\begin{table}[t!]
\centering
\caption{Comparison of different pseudo-label and program pairing strategies evaluated at the final iteration. Our proposed method, which uses paired synthetic programs and their executions for training, achieves the lowest Chamfer Distance and demonstrates the most consistent performance improvement across iterations.}
\label{tab:different_pairs}
\begin{tabularx}{\linewidth}{lc}
\toprule
\textbf{Sampling Method} & \textbf{Final Average CD} \\
\midrule
Our Method   & 9.73 \\
CAD-Recode   & 26.12 \\
Baseline 1 (best sample, its execution)  & 28.24 \\
Baseline 2 (best sample, input shape) & 10.28 \\
Baseline 3 (In Batch Sampling)  & 22.84 \\
\bottomrule
\end{tabularx}
\end{table}

\subsubsection{Baseline 1: (best sample, execution) pair}
\label{baseline_1}
This baseline trains on generated programs paired with their executions. Performance degrades because training data increasingly drifts from the target shape distribution.

\subsubsection{Baseline 2: (best sample, input shape) pair}
\label{baseline_2}
This baseline pairs generated programs with input shapes. It yields improvements but introduces supervision noise since programs do not exactly reconstruct the paired shapes.

\subsubsection{Baseline 3: In Batch Sampling}
\label{baseline_3}
This variant of Baseline 2 trains only on the top 20\% of samples per batch. While top-performing shapes improve, lower-performing cases receive no updates, limiting overall gains.

\section{Conclusion}
\label{sec:conclusion}
We presented PLLM, a data-centric self-training framework for CAD program synthesis that treats unlabeled shapes as a source for constructing synthetic supervision. By iteratively generating, curating, and enriching CAD programs, PLLM converts raw shape collections into a growing program–shape dataset that supports model improvement without requiring paired annotations. 

Our pipeline combines program sampling, execution-based curation, and program-level augmentation to expand both the quality and diversity of synthetic programs. Our evaluations demonstrate that this synthetic data generation process consistently improves geometric fidelity and program diversity, allowing PLLM to outperform the baseline CAD-Recode model with lower Chamfer Distances across iterations while maintaining valid and interpretable CAD code.

A major limitation of our approach is computational cost. Iterative data synthesis requires repeated sampling, execution, and filtering of programs. However, this cost reflects the trade-off of replacing manual annotation with automated supervision construction. As program executors and sampling efficiency improve, we expect such data-centric approaches to become increasingly practical for scaling CAD program learning to large unlabeled repositories.

\clearpage

{
\clearpage

    \small
    \bibliographystyle{ieeenat_fullname}
    \bibliography{main}
}


\end{document}